\pdfoutput=1

\documentclass[11pt]{article}

\usepackage[final]{acl}

\usepackage{times}
\usepackage{latexsym}

\usepackage[T1]{fontenc}

\usepackage[utf8]{inputenc}

\usepackage{microtype}

\usepackage{inconsolata}

\usepackage{graphicx}

\usepackage{booktabs}
\usepackage{CJKutf8}
\usepackage{multirow}
\usepackage{soul}
\usepackage{arydshln}
\usepackage{graphicx}
\usepackage{subcaption}
\usepackage[utf8]{inputenc}
\usepackage{silence}
\title{Drivel-ology: \\Challenging LLMs with Interpreting Nonsense with Depth}

 \author{
    \textbf{Yang Wang\textsuperscript{1}}\quad 
    \textbf{Chenghao Xiao\textsuperscript{2}}\quad 
    \textbf{Chia-Yi Hsiao\textsuperscript{2}}\quad \\
    \textbf{Zi Yan Chang\textsuperscript{3}}\quad 
    \textbf{Chi-Li Chen\textsuperscript{3}}\quad 
    \textbf{Tyler Loakman\textsuperscript{3}}\quad 
    \textbf{Chenghua Lin\textsuperscript{1}}\thanks{\, \, Corresponding author.} 
    \\
    \textsuperscript{1}The University of Manchester\quad
    \textsuperscript{2}Durham University\quad
    \textsuperscript{3}The University of Sheffield
    \\
    \texttt{yang.wang-27@postgrad.manchester.ac.uk}\quad
    \texttt{chenghua.lin@manchester.ac.uk}
 }

\begin{document}
\maketitle
\begin{abstract}
We introduce Drivelology, a unique linguistic phenomenon characterised as ``nonsense with depth'' -- utterances that are syntactically coherent yet pragmatically paradoxical, emotionally loaded, or rhetorically subversive. While such expressions may resemble surface-level nonsense, they encode implicit meaning requiring contextual inference, moral reasoning, or emotional interpretation. We find that current large language models (LLMs), despite excelling at many natural language processing (NLP) tasks, consistently fail to grasp the layered semantics of Drivelological text. To investigate this, we construct a benchmark dataset of over 1,200+ meticulously curated and diverse examples across English, Mandarin, Spanish, French, Japanese, and Korean. Each example underwent careful expert review to verify its Drivelological characteristics, involving multiple rounds of discussion and adjudication to address disagreements. Using this dataset, we evaluate a range of LLMs on classification, generation, and reasoning tasks. 
Our results reveal clear limitations of LLMs: models often confuse Drivelology with shallow nonsense, produce incoherent justifications, or miss implied rhetorical functions altogether. 
These findings highlight a deep representational gap in LLMs' pragmatic understanding and challenge the assumption that statistical fluency implies cognitive comprehension. We release our dataset
and code
to facilitate further research in modelling linguistic depth beyond surface-level coherence.
\end{abstract}


\begin{center}
\begin{tabular}{c@{\hskip 0.2cm}l}
    \raisebox{-.25\height}{\href{https://huggingface.co/datasets/extraordinarylab/drivel-hub}{\includegraphics[width=0.5cm]{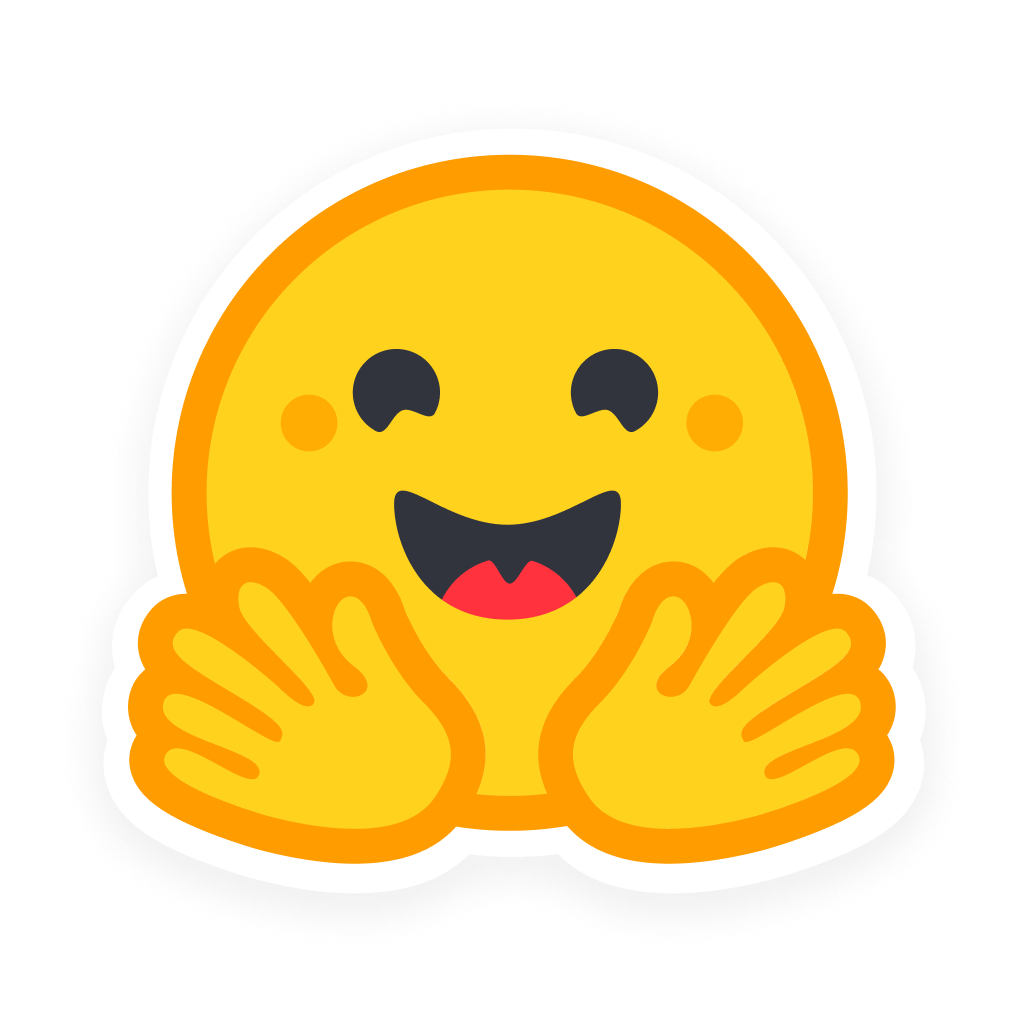}}} & {\small\texttt{\href{https://huggingface.co/datasets/extraordinarylab/drivel-hub}{extraordinarylab/drivel-hub}}}
\end{tabular}
\end{center}

\begin{center}
\begin{tabular}{c@{\hskip 0.2cm}l}
    \raisebox{-.25\height}{\href{https://github.com/ExtraOrdinaryLab/drivelology}{\includegraphics[width=0.4cm]{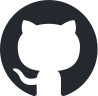}}} & {\small\texttt{\href{https://github.com/ExtraOrdinaryLab/drivelology}{github.com/ExtraOrdinaryLab/drivelology}}}
\end{tabular}
\end{center}

\section{Introduction}

Large language models (LLMs) have achieved impressive success across a wide range of natural language processing tasks, from machine translation and summarisation to commonsense reasoning and dialogue generation \cite{tang-etal-2023-enhancing,achiam2023gpt, qwen2.5, liu2024deepseek, guo2025deepseek, wang2025adversarial, goldsack-etal-2025-facts}. 
These models exhibit high degrees of fluency, contextual awareness, and even emergent reasoning capabilities. 
However, whether such performance reflects genuine understanding or merely statistical pattern-matching remains an open and pressing question \cite{bender2021dangers, rayhan2023artificial}.


The continuous evolution of Internet language as a distinct linguistic style offers a novel and insightful avenue for exploring the depth of understanding in LLMs \cite{10589379, mei2024slang}. 
Internet language, characterised by its dynamic evolution and cultural embedding, serves as an effective indicator to assess whether models truly grasp deeper semantics or simply rely on superficial pattern recognition. 
In particular, we introduce the term \textit{Drivelology}, combining \textit{drivel} (i.e., nonsense) with \textit{-ology} (i.e., the study of), which exemplifies this complexity. 
Drivelology often involves narratives with dual or multiple layers of meaning, employing non-linear structures and ambiguous expressions that challenge LLMs. 
Unlike purely nonsensical yet grammatically correct sentences such as ``\textit{Colourless green ideas sleep furiously}'' \cite{Chomsky57a}, or simplistic tautologies like ``\textit{either it is or it isn't}'',  Drivelology intentionally embeds subtle cultural references, irony, or satire within superficially trivial or absurd narratives. 
For example, ``\textit{I deeply admire Che Guevara’s anti-capitalist spirit, so I bought all his merchandise}'' illustrates how Drivelology paradoxically critiques performative activism but still grammatically sound. 
Thus, it differs from typical internet content, such as inspirational quotes or prose, by demanding deeper interpretative engagement from both human readers and LLMs.


Existing studies have explored the difficulty for LLMs to understand humour, sarcasm, and irony \cite{loakman-etal-2023-iron,loakman2025comparingapplesorangesdataset,romanowski2025punchlines, zheng2025fanchuan}. 
However, Drivelology differs fundamentally from these phenomena by employing more complex narratives and deeper ambiguities, making it a uniquely challenging benchmark for assessing LLMs' semantic comprehension. 
Studying LLMs' ability to handle Drivelology offers insights into their social and semantic reasoning, as it encodes subtle emotions and culturally embedded meanings. 
Understanding such linguistic forms is essential for developing socially intelligent systems \cite{gandhi2023understanding, kosinski2024evaluating, mittelstadt2024large}. 
Moreover, enhancing AI's grasp of Drivelology can potentially boost creativity in AI applications, improving user experience in content creation tools \cite{hu2024cracking}, and even contribute to model safety \cite{matamoros2023humour} through better understanding of contextually ambiguous content.

Our contributions are as follows:
\begin{itemize}
    \item We design a novel taxonomy for Drivelological narratives to aid in categorising the source of meaning embedded in the text.
    \item We collect and rigorously annotate a novel benchmark dataset called \textsc{DrivelHub} for understanding Drivelology from the internet, consisting of 1,200+ Drivelological examples that are considered \textit{nonsense with depth}. 
    \item We use \textsc{DrivelHub} as the basis for four novel tasks: (1) Drivelology Detection: A binary classification task to determine whether a given text is Drivelology or non-Drivelology; (2) Drivelology Tagging: A multi-label classification task to assign one or more categories from our taxonomy (\S\ref{sec:drivelology}) to Drivelology samples; (3) Implicit Narrative Writing: An implicit narrative explanation task for a given Drivelology sample; and (4) Narrative Selection: A multiple-choice task where the model selects the correct narrative from five options. 
\end{itemize}

These tasks collectively encompass various levels of Drivelology understanding, ranging from literal content comprehension to more sophisticated narrative reasoning, thereby providing a comprehensive assessment of Drivelology understanding capabilities. 
We conducted extensive experiments using the \textsc{DrivelHub} dataset, evaluating both proprietary and open-source LLMs. 







\section{Related Work}
\label{sec:related_works}

\noindent\textbf{LLMs Evaluations.}~~Recent LLMs have demonstrated remarkable performance in following human instructions and performing various downstream tasks
through zero-shot prompting \cite{naveed2023comprehensive, liang2024survey, chang2024survey, liu2024mmbench, liu-etal-2024-llms-narcissistic}. 
Various benchmarks have been proposed to evaluate their performance, primarily focusing on assessing the fundamental capabilities of LLMs \cite{zellers-etal-2019-hellaswag, sakaguchi2021winogrande, hendrycks2021measuring, suzgun-etal-2023-challenging, zheng2023judging, zhou2023instructionfollowingevaluationlargelanguage, jimenez2024swebench, yang2024swebenchmultimodal, wang-lin-2025-tougher, hong2025beyond}. 
However, the ability of large models to perform in-depth social reasoning and accurately understand human contexts remains underexplored \cite{hu2023language, feng2024how}. 


\noindent\textbf{Humour, Irony, and Sarcasm.}~~Humour, irony, and sarcasm are fundamental elements of human interaction, each requiring a deep understanding of language, context, and social cues \cite{palmer2003taking, filik2016sarcasm, koder2021irony, mir2021spanish, loakman-etal-2023-iron, loakman2025comparingapplesorangesdataset}. While computational approaches have addressed these phenomena \cite{yang2015humor, jentzsch2023chatgpt, boutsikaris2024comparative}, they often focus on resolving a central contradiction between a literal statement and its context \cite{kreuz2002asymmetries, misra2019sarcasm}. 
Classic sarcasm, for example, is typically understood through a single cognitive step: inverting a word's meaning based on a negative premise, as in, ``\textit{Forgetting assignments and stressing over grades, what a fun semester}''.

We argue that Drivelology presents a more profound challenge, distinguished by two key characteristics: \textbf{(1) its compositional, multi-layered structure, and (2) its use of pragmatic paradox and ambiguity.} 
For example, ``\textit{I deeply admire Che Guevara's anti-capitalist spirit, so I bought all his merchandise}'' is not a simple semantic inversion. Its critique of performative activism requires synthesising cultural knowledge, making irony just one component of its layered meaning. Furthermore, Drivelology uses pragmatic paradoxes like ``\textit{I'm good at everything except what I can't do.}'' This statement is not explicitly sarcastic nor ironic; its challenge lies in navigating the ambiguity of the speaker's intent. This reliance on compositional meaning and deliberate ambiguity is what sets Drivelology apart.

\begin{figure*}[!t]
    \centering
    \includegraphics[width=0.9\textwidth]{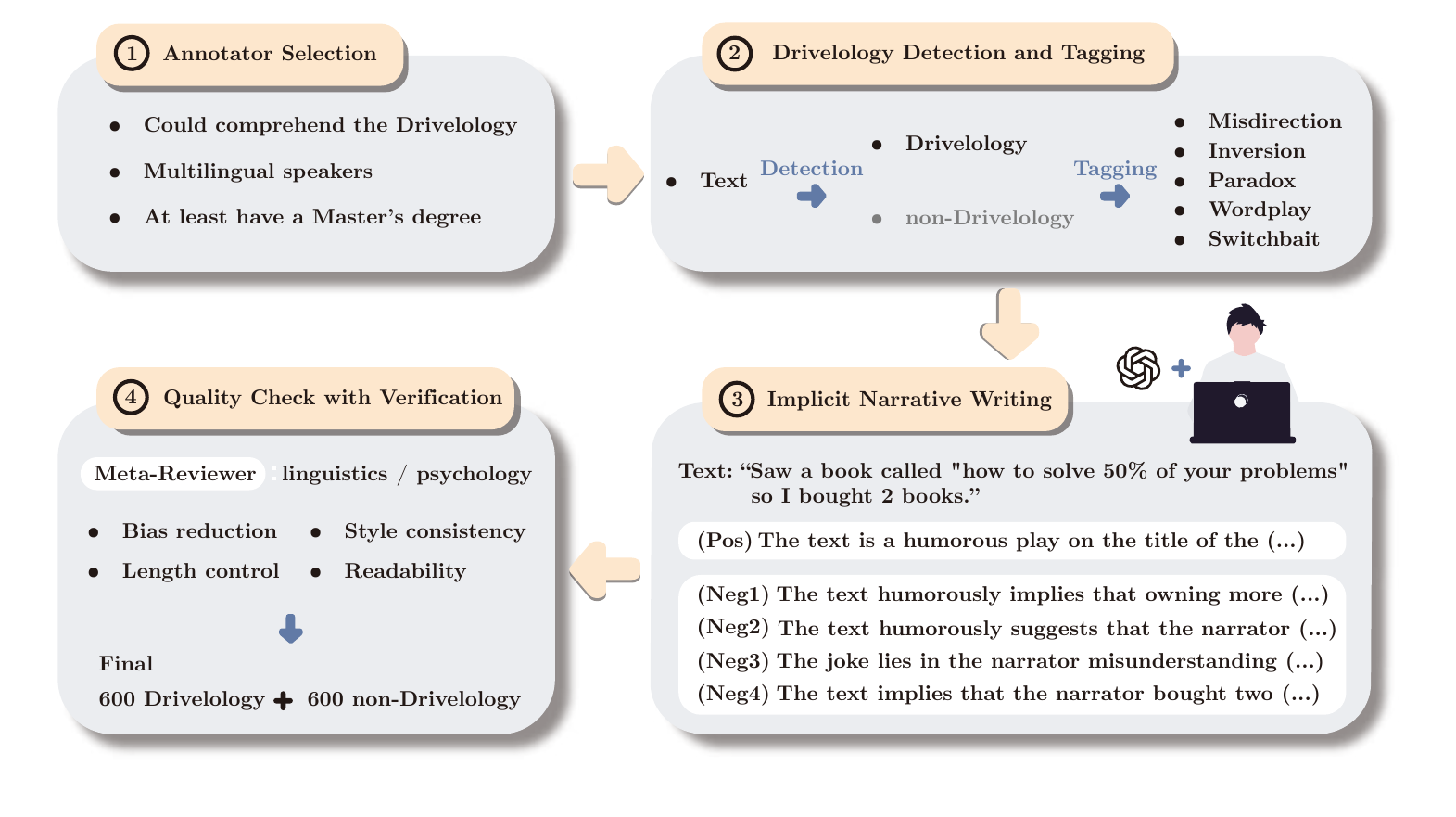}
    \caption{Overview of the multi-stage process for constructing the \textsc{DrivelHub} dataset.}
    \label{fig:overview_dataset_construction}
\end{figure*}

\noindent\textbf{Distinguishing Drivelology from Deceptive and Nonsensical Language.}~~To properly situate our work, it is crucial to distinguish Drivelology from related pragmatic concepts. 
\citet{cappelen2019bad} identify a category termed \textit{deep bullshit}: utterances defined by an indifference to whether the words make any sense at all, resulting in genuine nonsense. 
For example, a statement like ``\textit{Colourless green ideas sleep furiously}'' \cite{Chomsky57a} qualifies as deep bullshit as it is semantically null. 
This is distinct from the more widely discussed Frankfurt-style \textit{bullshit}, which is characterised by an indifference to \textit{truth} rather than meaning, often deployed to persuade without regard for fact \cite{frankfurt_bullshit_2005}. 
For instance, a politician might declare they bring a ``\textit{fresh perspective, unburdened by the stagnant thinking of Washington insiders}'', a statement chosen for its persuasive effect, not its accuracy. 
Drivelology shares a superficial resemblance with deep bullshit, as both can appear nonsensical. However, the two are fundamentally antithetical in their purpose and construction. Whereas deep bullshit arises from a \textit{disregard for meaning}, Drivelology is meticulously crafted \textit{for the sake of conveying a hidden meaning}. It is, as we define it, ``nonsense with depth''. 

The surface-level absurdity of a Drivelological text is a deliberate rhetorical framework, designed to guide an audience toward an implicit critique, observation, or emotional payload. Thus, unlike vacuous deep bullshit, Drivelology is rhetorically complex and purposeful. 
While other forms of bad language, such as lying or misleading, are defined by their deceptive relationship to truth \cite{cappelen2019bad}, Drivelology’s defining feature is its purposeful and creative use of apparent nonsense to generate layered semantics. Clarifying these boundaries is essential for developing AI systems that can appreciate subtle human expression. A truly capable model must differentiate between genuinely meaningless utterances (deep bullshit) and the sophisticated, implicit communication of Drivelology, a task that requires moving beyond surface-level coherence to grasp complex rhetorical intent.

\section{The \textsc{DrivelHub} Dataset}

\begin{figure*}[!t]
    \centering
    \includegraphics[width=1.0\textwidth]{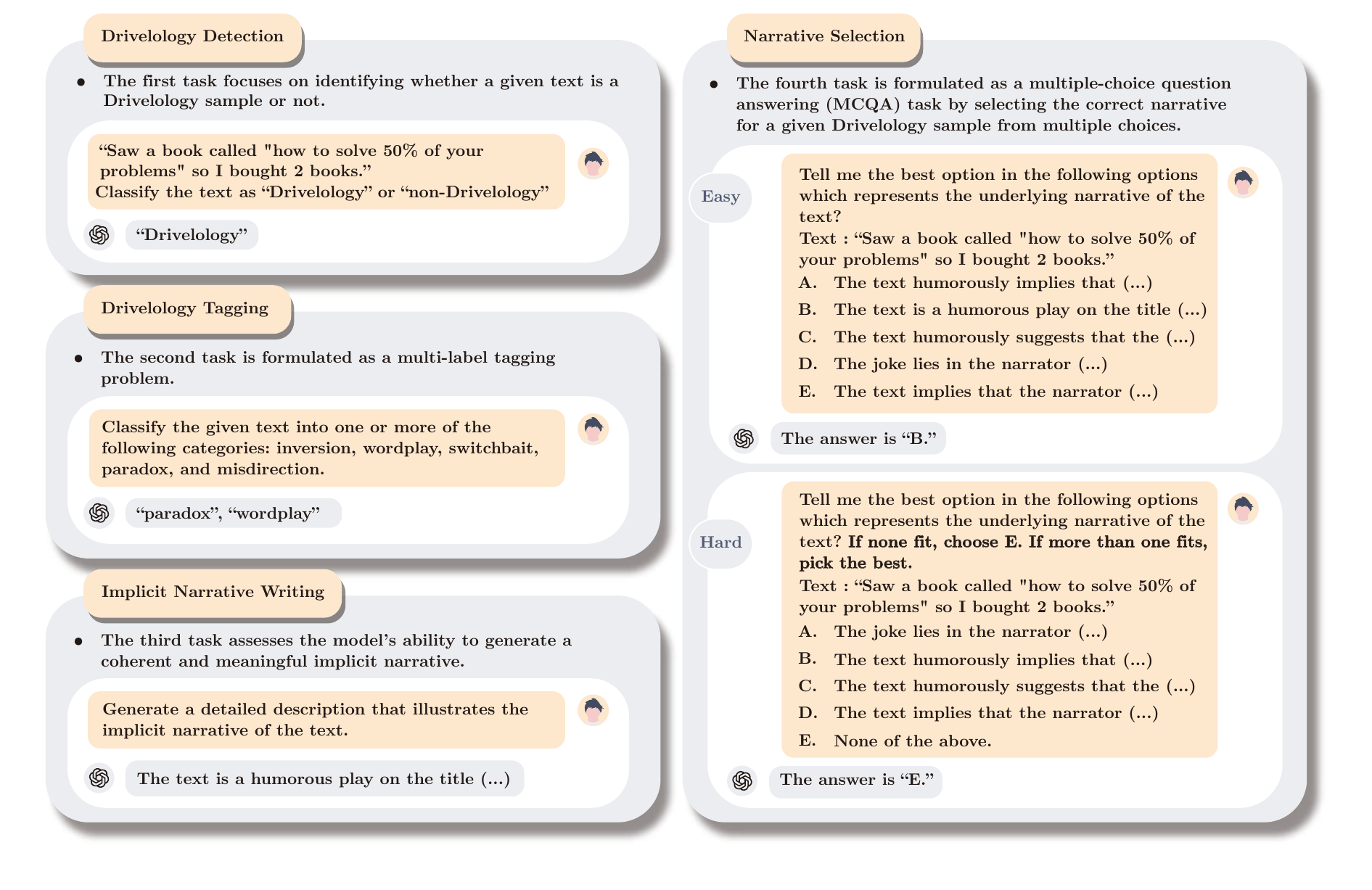}
    \caption{Overview of the Drivelology evaluation framework for LLMs. The figure illustrates four core tasks designed to systematically assess LLMs' ability to understand and reason about Drivelology: Drivelology Detection (binary classification), Drivelology Tagging (multi-label classification), Implicit Narrative Writing (generative reasoning), and Narrative Selection (multiple-choice question answering with both Easy and Hard settings).}
    \label{fig:task_overview}
\end{figure*}

Our benchmark dataset, \textsc{DrivelHub}, is designed to evaluate how well LLMs understand Drivelology. 
Each entry in the dataset includes: (1) a Drivelology sample, (2) the underlying message that the sample aims to convey, and (3) one or more categories describing the main type of Drivelology contained in the sample. 
These components form the basis for a variety of tasks that assess different aspects of Drivelology comprehension and reasoning. 
An overview diagram of the multi-stage process for constructing the \textsc{DrivelHub} dataset is presented in Figure~\ref{fig:overview_dataset_construction} and Appendix~\ref{sec:overview_annotation}.

\subsection{Taxonomy of Drivelology}
\label{sec:drivelology}

Drivelology refers to a unique style of language that blends humour, ambiguity, and rhetorical complexity to create statements that are intentionally puzzling or nonsensical. 
Unlike ordinary nonsense or straightforward jokes, Drivelology often relies on layered meanings, unexpected twists, and linguistic playfulness to engage readers in deeper interpretation or amusement. 
The defining characteristics of Drivelology can be broadly categorised into the following taxonomy:


\noindent\textbf{Misdirection.}~~This technique leads the listener down an expected path before a final twist reveals a different, often more literal or absurd, ending. 
Example: \textit{``Don’t give up on your dream so easily! Keep sleeping!''} The expected path is motivational encouragement; the twist is a literal interpretation of ``dream.''


\noindent\textbf{Paradox.}~~This relies on a statement that appears logically self-contradictory but contains a latent, often humorous or profound truth. The core of the technique is the clash of seemingly incompatible ideas. 
Example: \textit{``I will not forget this favour until I forget it.''} This is a logically circular statement that humorously asserts the certainty of remembering.


\noindent\textbf{Switchbait.}~~
This technique hinges on a specific phrase (the ``bait'') that has a culturally-embedded double meaning. The initial context is then suddenly replaced (the ``switch'') by a surprising second meaning. The humour is generated by this cynical or culturally-specific reinterpretation of the bait, rather than by derailing a narrative. 
Example: \textit{``Brit: You've got a gun problem. American: Yeah, at least it's a modern problem.''} The bait is the phrase ``gun problem.'' The switch reframes it from a criticism of US gun violence to a dark turn, implying cultural counter-attack on UK knife crime.


\noindent\textbf{Inversion.}~~This technique takes a well-known phrase, cliché, or social script and flips it on its head. The humour arises by reversing a familiar structure to creating a new, often satirical, meaning. 
Example: \textit{``Other than being good-looking, having a great figure, and having money, I have nothing else.''} This inverts the structure of a humble complaint into an arrogant boast.


\noindent\textbf{Wordplay.}~~This is the use of linguistic creativity, often by exploiting the phonetics or polysemy of words. It includes puns, double entendres, and similarities. 
Example: \textit{``Do you have any raisins? No? How about a date?''} This is a classic homographic pun playing on two meanings of the word ``date''.

We note that the defining characteristic of Drivelology is not the use of a single technique, but the creative and often simultaneous combination of several within a single utterance to produce its layered, nonsensical effect. This inherent complexity is central to our study, which is why the Drivelology Tagging task is formulated as a multi-label classification problem (see \S\ref{sec:label-design}), allowing a single sample to be annotated with one or more of the following categories.

\subsection{Drivelology Collection}


To ensure a comprehensive evaluation, we prioritised high diversity in our benchmark dataset by selecting a wide range of topics from different languages. 
Our data was collected from a variety of popular platforms (e.g., Instagram, Threads, TikTok, Facebook, Line, RedNote, Pinterest, Naver, and YouTube). 
These platforms were chosen strategically as their largest user demographic falls between 25 to 34 years old,\footnote{According to Statista's social media demographics data as of September 2025. \url{https://www.statista.com/topics/1164/social-networks}} which aligns well with our research focus since Drivelology content predominantly originates from younger generations \cite{Sharula2024}. 
For non-Drivelology samples, we curated content from sources such as famous quotes, proverbs, and Ruozhiba (a popular online forum). 
These non-Drivelology samples are also multilingual, covering English, Mandarin, Spanish, French, Japanese, and Korean. 
We further categorised non-Drivelology samples into two types: normal sentences (such as meaningful quotes or proverbs) and pure nonsense (text that lacks logical structure or meaning). 
A significant proportion of the pure nonsense samples were collected from Ruozhiba.

\subsection{Data Annotation}
\label{sec:data_annotation}

Labelling Drivelology requires both content comprehension and cultural context understanding. 
We implemented a rigorous four-step annotation protocol: 
(1) \textbf{Annotator Selection:}~~We assembled a team of seven multilingual annotators who all held at least a Master's degree and demonstrated proficiency in multiple languages.\footnote{The annotation team consisted of four authors of this paper and three paid annotators recruited for their linguistic expertise.} 
(2) \textbf{Drivelology Detection and Tagging:}~~Annotators identified texts as either Drivelology or non-Drivelology, and classified Drivelology samples into categories including Misdirection, Paradox, Switchbait, Inversion, and Wordplay. 
(3) \textbf{Implicit Narrative Writing:}~~We employed a human-in-the-loop process to create the narrative explanations. For each Drivelology sample, human experts drafted and refined the correct narrative explanation. We then utilised GPT-4.5\footnote{gpt-4.5-preview-2025-02-27} as an assistive tool to generate four plausible but incorrect narrative counterparts, all of which underwent a final stage of manual verification and editing to ensure their quality as effective distractors. 
(4) \textbf{Quality Check:}~~A meta-reviewer with linguistics and psychology expertise reviewed all annotations. 
The meta-reviewer also revised the narratives as needed to ensure consistent length, uniform writing style, and improve overall readability. 
Further details concerning the annotation process can be found in Appendix~\ref{sec:overview_annotation}.

\subsection{Task Design}
\label{sec:label-design}

To evaluate an LLM's ability to understand Drivelology, we designed four tasks to assess different facets of social and non-linear reasoning. An overview of these tasks is provided in Figure~\ref{fig:task_overview}.

\noindent\textbf{Drivelology Detection.}~~
A binary classification task where the model must determine if a given text is Drivelology or non-Drivelology.

\noindent\textbf{Drivelology Tagging.}~~
A multi-label classification task where the model assigns one or more descriptive categories (see \S\ref{sec:drivelology}) to a Drivelology sample to capture its layered rhetorical structure.

\noindent\textbf{Narrative Writing.}~~
A generative task where the model explains the implicit narrative and underlying meaning of a Drivelology sample, requiring it to move beyond a surface-level reading.

\noindent\textbf{Narrative Selection.}~~
A multiple-choice question answering (MCQA) task where the model is tasked with selecting the correct narrative for a Drivelology sample from several options. The \textbf{Easy} version offers one correct answer and four distractor, whilst the \textbf{Hard} version adds a ``none of the above'' option, requiring deeper reasoning, as this option should only be chosen if none of the provided narratives adequately capture the underlying meaning of the Drivelology sample. 
This additional step significantly increases the task's complexity, as it prevents reliance on simple elimination strategies.

\section{Experiments}

\subsection{Models and Settings}

We evaluate the performance of state-of-the-art LLMs in a zero-shot setting. 
We utilise both proprietary models including GPT-4 \cite{achiam2023gpt} and Claude-3 \cite{anthropic2024claude}, as well as open-sourced models including Qwen3 \cite{yang2025qwen3}, Qwen2.5 \cite{qwen2.5}, Llama3.1 \cite{grattafiori2024llama}, Llama3 \cite{grattafiori2024llama}, and DeepSeek V3 \cite{liu2024deepseek}.

To minimise variance across task prompts, we design three distinct prompts for each task and report the average performance over three runs (one for each prompt). 
Detailed descriptions of the prompts and additional experimental settings are provided in Appendix~\ref{sec:experiment_prompts}.

\subsection{Evaluation Metrics}

We use accuracy for the Drivelology Detection task, F1 score for the Drivelology Tagging task, and accuracy for the MCQA task. 
For the generation task that involves writing narrative explanations, we apply reference-based evaluation metrics commonly used in text generation studies \cite{celikyilmaz2020evaluation}.
Specifically, we use BERTScore \cite{Zhang2020BERTScore} and an LLM-as-a-judge evaluation paradigm \cite{zheng2023judging}. 
Recent work shows that GPT-based evaluation aligns well with human judgments \cite{chan2023clair, liu2023g, hu2024cracking, gu2024survey}, and thus we select GPT-4 series for LLM-as-a-judge evaluation. 
The judge was tasked to rate each generated narrative on a 1-to-5 Likert scale based on its semantic quality. 
Note that we use different GPT variants for different purposes: gpt-4.5 for data annotation, gpt-4o-mini for zero-shot experiments, and gpt-4.1 for LLM-as-a-judge evaluation in text generation tasks. This helps reduce potential evaluation bias toward GPT-4's own generation \cite{hu2024cracking, liu-etal-2024-llms-narcissistic}.

\section{Main Results}
\label{sec:main}

\begin{table*}[!t]
\centering
\begin{tabular}{lrrrrrr}\toprule
\multirow{2}{*}{\textbf{Models}} &\multicolumn{2}{c}{\textbf{Narrative}} &\multicolumn{2}{c}{\textbf{MCQA}} &\multicolumn{2}{c}{\textbf{Classification}} \\\cmidrule{2-7}
&\textbf{BERT} &\textbf{GPT} &\textbf{Easy} &\textbf{Hard} &\textbf{Detect} &\textbf{Tag} \\\midrule
gpt-4o-mini &85.81 &2.90 &81.89 &4.67 &\ul{75.00} &49.52 \\
claude-3.5-haiku &\ul{86.51} &\ul{3.39} &\ul{83.17} &11.56 &71.90 &\ul{52.03} \\
\midrule
llama-3-8b-instruct &84.67 &2.63 &77.39 &1.67 &57.81 &39.90 \\
llama-3.1-8b-instruct &85.60 &2.75 &77.56 &1.89 &58.57 &36.21 \\
qwen2.5-7b-instruct &85.51 &2.78 &77.50 &3.78 &62.66 &42.49 \\
qwen3-8b-instruct &85.91 &2.64 &\ul{83.17} &\textbf{26.78} &65.00 &38.04 \\
deepseek-v3 &\textbf{87.11} &\textbf{3.59} &\textbf{86.83} &\ul{15.50} &\textbf{81.67} &\textbf{55.32} \\
\bottomrule
\end{tabular}
\caption{Main results. For the narrative explanation writing task, we report BERTScore-recall (BERT) and GPT-4-as-a-judge (GPT) evaluation scores. For the narrative selection task, we report accuracy. For the Drivelology classification tasks, we report accuracy for the detection task and weighted F1 score for the tagging task. The best scores are \textbf{bold} and the second best ones are \ul{underlined}.}\label{tab:main}
\end{table*}

The main results in Table~\ref{tab:main} show a clear hierarchy in model performance. Deepseek-v3 is the dominant model, achieving the top score in five of the six evaluated metrics. The contrast between the two evaluation metrics in the Narrative Writing task is particularly noteworthy. 
While BERTScore-recall values are high across all models, suggesting a universal proficiency in generating fluent text, the GPT-4-as-a-judge scores provide a much clearer picture of true narrative quality. On this scale, deepseek-v3 (3.59) and claude-3.5-haiku (3.39) are the only models to score comfortably above three, indicating their outputs were judged as possessing high semantic quality. In stark contrast, other models like llama-3-8b-instruct (2.63) and qwen3-8b-instruct (2.64) fall below this threshold, suggesting their narratives failed to capture the required depth and were deemed qualitatively weaker by the LLM-as-a-judge. 
The most striking performance gap is present in the MCQA task. 
The Hard setting causes a steep decline in accuracy for all models, exposing a critical weakness in subtle reasoning. Notably, qwen3-8b-instruct is an outlier here, scoring 26.78\%, which far surpasses the next-best model. 
In the Classification tasks, deepseek-v3 again confirms its superior understanding by leading in both the Detection (81.67\%) and Tagging (55.32\%) tasks.

\begin{figure*}[!t]
    \centering
    \includegraphics[width=1.0\textwidth]{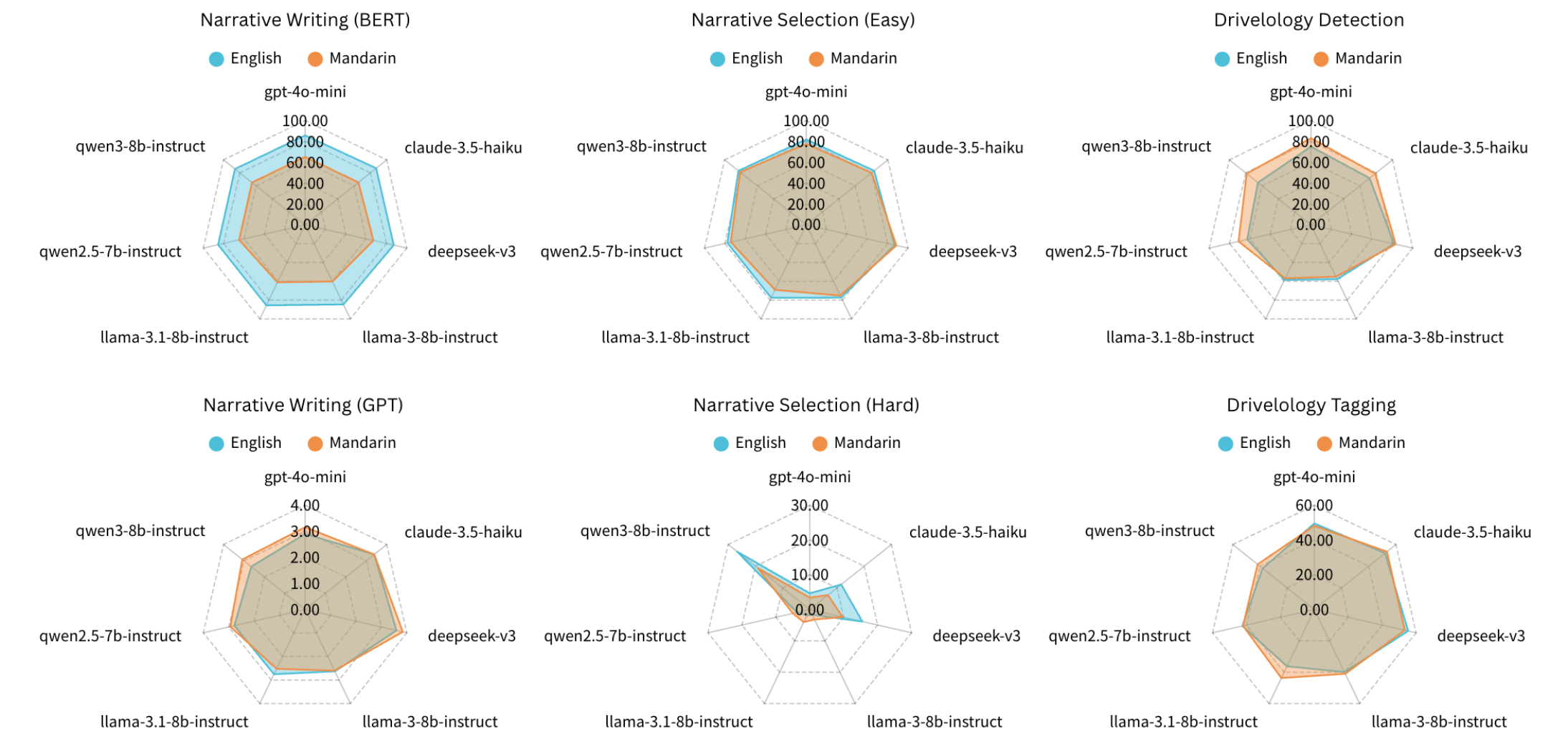}
    \caption{Model performance on the multilingual \textsc{DrivelHub} dataset, contrasted by prompt language (English vs. Mandarin). Each reported score is the average performance over three distinct prompts to minimise variance.}
\label{fig:radar}
\end{figure*}

\subsection{Prompt Language Influence}

An analysis of the impact of prompt language on model performance reveals a metric-dependent pattern. As shown in Figure~\ref{fig:radar}, the choice between English and Mandarin prompts is not a neutral choice but a significant factor that influences evaluation outcomes. We identify two distinct and opposing patterns. 
Firstly, English prompts consistently yield superior performance on tasks that reward lexical precision and complex logical reasoning. In the Narrative Writing task, this is most evident in the BERTScore results, where every model scores higher when prompted in English. This suggests that English instructions may prime the models to generate outputs with greater lexical overlap with the reference translations, a feature to which BERTScore is highly sensitive \cite{hanna2021fine}. A similar advantage for English prompts is observed in the MCQA task. The uniform improvement in both Easy and Hard settings implies that English may serve as a more robust internal \textit{language of reasoning}. 
Conversely, Mandarin prompts produce a consistent, albeit smaller, advantage on tasks that prioritise direct content comprehension. 
The improved performance under GPT-as-a-judge, which evaluates qualitative coherence, indicates that Mandarin prompts better align the models with the semantic and narrative intent of the source material. The consistent gains in the classification tasks further suggest that for direct comprehension and categorisation, instructions in Mandarin are more effective.

\subsection{Model Size Scaling in the Qwen3 Series}


The results in Table~\ref{tab:qwen_scaling_v2} illustrate the impact of model size on performance. We focus on the Qwen3 series across the MCQA and Classification tasks, as these tasks exhibited the widest performance variance in Table~\ref{tab:main}, making them most suitable for studying scaling effects. 
For the Easy MCQA task, performance gains are consistent but modest: as model size increases from 4B to 14B, accuracy improves by approximately 3\% for English prompts and 6\% for Mandarin. The Hard task reveals a spiking scaling effect. With English prompts, accuracy leaps from a mere 6.00\% for the 4B model to 45.83\% for the 14B model. The trend is even more pronounced with Mandarin prompts, where the score skyrockets from 2.44\% to 47.89\%. This indicates that the more complex reasoning required by the Hard task is a key differentiator that is unlocked by larger model sizes. Therefore, the ability to handle such complex reasoning appears to be an emergent property in the Qwen3 architecture, strongly correlated with its parameter count. 
For the Detection task, performance does not consistently improve with size. Notably, when prompted in Mandarin, the 8B model significantly outperforms both its smaller and larger counterparts. The Tagging task reveals yet another pattern: a noticeable dip in performance at the 8B size, which then recovers to achieve the best score at 14B for both languages. 
These findings indicate that the benefits of model scaling are task-dependent.

\begin{table}[!t]
\centering
\resizebox{\linewidth}{!}{
\begin{tabular}{lcrrrr}
\toprule
\multirow{2}{*}{\textbf{Prompt}} & \multirow{2}{*}{\textbf{Size}} & \multicolumn{2}{c}{\textbf{MCQA}} & \multicolumn{2}{c}{\textbf{Classification}} \\
\cmidrule(lr){3-6}
& & \textbf{Easy} & \textbf{Hard} & \textbf{Detect} & \textbf{Tag} \\
\midrule
\multirow{3}{*}{English} & 4B & 81.00 & 6.00 & 66.80 & 43.21 \\
& 8B & 83.17 & 26.78 & 65.00 & 38.04 \\
& 14B & 83.94 & 45.83 & 66.22 & 47.61 \\
\midrule
\multirow{3}{*}{Mandarin} & 4B & 77.61 & 2.44 & 62.86 & 46.10 \\
& 8B & 81.11 & 19.11 & 78.81 & 41.71 \\
& 14B & 83.50 & 47.89 & 71.78 & 49.13 \\
\bottomrule
\end{tabular}
}
\caption{MCQA and Classification results in the Qwen3 series of different sizes. This table shows the performance on both tasks when prompted in English and Mandarin. The Full version containing all tasks can be found in Table~\ref{tab:qwen_scaling_full}.}
\label{tab:qwen_scaling_v2}
\end{table}

\subsection{Role of Language in the MCQA Task}

\begin{figure*}[!t]
    \centering
    \includegraphics[width=1.0\textwidth]{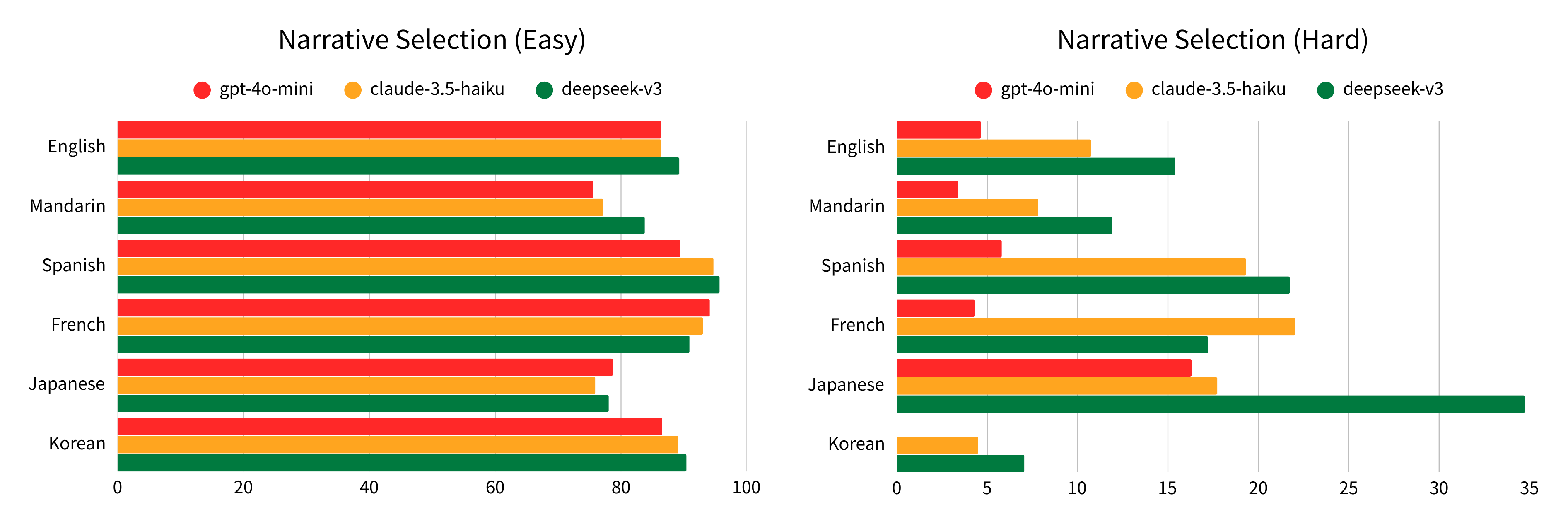}
    \caption{A language-based breakdown of Narrative Selection (MCQA) accuracy from Table~\ref{tab:main}. The charts disaggregate the overall Easy and Hard accuracy scores based on the original language of the Drivelology sample.}
\label{fig:mcqa_language}
\end{figure*}

A closer look at the MCQA results from Table~\ref{tab:main} reveals that aggregate scores mask significant performance variations across the different languages in \textsc{DrivelHub}. As shown in the breakdown in Figure~\ref{fig:mcqa_language}, we can analyse the difficulty of each language's content for the models. 
Deepseek-v3 consistently demonstrates the most robust cross-lingual performance, achieving the highest accuracy across nearly all languages in both the Easy and Hard settings. The analysis also pinpoints which languages pose a greater challenge. Korean and Mandarin consistently result in the lowest accuracy, especially in the Hard task, marking their content as the most difficult for the models to process. 

\section{Analysis and Discussion}

\subsection{Analysis of Model Reasoning}

In the Narrative Writing task, claude-3.5-haiku and deepseek-v3 achieve the highest GPT-4-as-a-judge scores (3.39 and 3.59) and also perform strongly in the Drivelology Tagging task (52.03\% and 55.32\%). 
Thie correlation between their performance in both tasks raises an important question: \textit{Do these models arrive at correct Drivelology classifications through appropriate reasoning that reflects a true understanding of the underlying meaning?} To investigate this question, we analyse their reasoning processes across several representative examples.

For example: ``\textit{Meng Po: Those who have forgotten their names, please follow me.}'' Deepseek-v3 categorise this as \textit{switchbait}, emphasising the cultural significance of Meng Po, a mythological figure who administers the Soup of Forgetfulness in Chinese folklore. 
Its reasoning explicitly highlights the importance of cultural knowledge, suggesting they treat Meng Po's mythological role as important context that readers must understand to appreciate the Drivelology. 
In contrast, claude-3.5-haiku categorises it as a \textit{paradox}, focusing on the logically self-contradictory statement: ``how can someone who has forgotten their name respond to such a call?'' This divergence in reasoning approaches suggests varying degrees of cultural knowledge \textit{internalisation} among models. 
Claude-3.5-haiku appears to have so thoroughly internalised the cultural context of Meng Po that it treats it as implicit knowledge, allowing it to focus on the logical structure of the text rather than its cultural elements. 
This observation raises important questions about how different models process and prioritise cultural knowledge versus logical reasoning in their analysis of culturally-embedded texts, and how such internalisation affects their ability to identify different categories of Drivelology.

\subsection{Analysis of Human Reasoning}

Drivelology challenges not only LLMs but also human annotators, who often bring diverse perspectives and interpretations to the same text. Because Drivelology is intentionally ambiguous, contradictory, or ironic, it invites multiple plausible readings. Annotators rely on their own linguistic, cultural, and contextual knowledge, which means that the same Drivelology sample can evoke different analytical frameworks depending on who is interpreting it.

Consider the statement: ``\textit{I hate two kinds of people: the first kind is those who don't finish their...}'' 
From a \textit{paradox} perspective, the speaker claims to dislike people who speak incompletely, yet the sentence itself is left unfinished, ironically exemplifying the very behaviour it criticises. This self-contradiction highlights the speaker's insincerity and creates a paradoxical effect. 
Alternatively, from a \textit{misdirection} viewpoint, the statement sets up the expectation of a complete list, but then abruptly stops, leaving the audience anticipating an answer that never comes. The humour and irony arise from this unresolved expectation. 
Another example is: ``\textit{I deeply admire Che Guevara's anti-capitalist spirit, so I bought all his merchandise.}'' 
Here, the \textit{paradox} lies in admiring Guevara's anti-capitalist stance while simultaneously engaging in capitalist consumerism by buying his merchandise. This contradiction turns ideological admiration into commercial participation. 
The \textit{switchbait} interpretation depends on cultural knowledge: recognising Che Guevara as a symbol of anti-capitalism is key. Without this context, the contradiction, and the humour, may not be apparent. 
The text's irony and layered meaning rely on shared cultural and historical understanding, making switchbait also an appropriate label.

\section{Conclusion}
\label{sec:conclusion}


In this work, we introduced Drivelology, a unique linguistic phenomenon that challenges the semantic and pragmatic understanding of LLMs. 
We constructed and evaluated the \textsc{DrivelHub} dataset across multiple languages and task settings. 
Our extensive experiments reveal a critical and consistent gap between statistical fluency and genuine comprehension in state-of-the-art LLMs. While models can generate syntactically coherent text, they largely fail to grasp the layered, culturally-embedded meanings central to Drivelology. We found that complex reasoning, particularly on the ``Hard'' MCQA task, remains a significant bottleneck, though performance scales predictably with model size. Conversely, performance on classification tasks showed non-linear scaling, suggesting that simply increasing parameter count is not a panacea for all reasoning deficits. 
The failure of LLMs to interpret Drivelology underscores a deep representational gap in their ability to model complex social and cultural contexts. Our work provides a concrete benchmark for the community to address these deeper challenges. Future research should focus not only on scaling models but also on developing novel training paradigms that explicitly target the multi-layered reasoning that defines sophisticated human communication.

\section*{Limitations}

\noindent\textbf{Language Imbalance.}~~Over one third of the samples in the \textsc{DrivelHub} dataset are in Mandarin (see Table~\ref{tab:distribution}). This results in a slight language imbalance, which may affect the generalisability of our findings across different linguistic and cultural contexts. 
We do our best in \S\ref{sec:main} to control for potential content distribution bias arising from this imbalance. 
Additionally, as the \textsc{DrivelHub} dataset is still being expanded, we will continue to focus on addressing distribution differences by increasing the representation of Drivelology samples in underrepresented languages.

\noindent\textbf{Limited Computation Resources.}~~Due to budget constraints, we were unable to evaluate stronger proprietary LLMs such as GPT-5, Claude-3.7, or DeepSeek R1, as their usage costs are prohibitively high. 
For open-source models, we restricted our experiments to 14B parameter models because of limited computational resources, and were unable to run larger models within our available infrastructure. 
We encourage researchers and the broader community to expand on this work by evaluating larger or more advanced LLMs as resources permit.

\noindent\textbf{Focus on Understanding Rather Than Generation.}~~In this paper, we focus on evaluating the understanding and reasoning abilities of LLMs with respect to Drivelology, rather than their capacity to generate fluent and human-like Drivelology text. While generation is an important aspect, it falls outside the main scope of our study. Nevertheless, we include a discussion in Appendix~\ref{sec:drivelology_generation} with sample generations, illustrating that current LLMs often require over 20 attempts to produce Drivelology that achieves comprehensive alignment between topic, rhetorical category, and sentence structure.

\section*{Ethics Statement}

\noindent\textbf{Copyright and License.}~~All data samples used in this study are collected exclusively from publicly available content on social media platforms. We respect the intellectual property rights of original authors by ensuring that no proprietary or paywalled material is included. The dataset is released solely for research purposes under a license that prohibits commercial use and redistribution of original content.

\noindent\textbf{Content Review and Harm Mitigation.}~~To uphold ethical standards, we carefully review all collected samples and filter out any content that may be offensive, harmful, or violate privacy. Our annotation process is designed to ensure that sensitive information is excluded and that the dataset does not propagate hate speech, harassment, or other forms of harmful language.

\noindent\textbf{Intended Use.}~~The dataset and accompanying resources are intended strictly for academic research and the advancement of natural language processing technologies. Users are advised to adhere to ethical guidelines and local regulations when using the dataset.

\section*{Acknowledgments}
Tyler Loakman is supported by the Centre for Doctoral Training in Speech and Language Technologies (SLT) and their Applications funded by UK Research and Innovation (EP/S023062/1).

\bibliography{custom}

\appendix


\clearpage

\section{Dataset Details}

\subsection{Overview of the Annotation Process}
\label{sec:overview_annotation}

Labelling Drivelology presents significant challenges, not only because it demands a deep familiarity with both content and cultural context, but also due to the potential for divergent interpretations among annotators from varied backgrounds. 
For each Drivelology sample, we annotate the underlying narrative and the category of the Drivelology. 
To ensure high-quality and precise annotations, we designed a multi-step annotation protocol as follows:
(1) \textbf{Annotator Selection.}~~We recruited multilingual annotators, and ensured that they could comprehend the Drivelology. 
Eight human judges\footnote{The original annotation was performed by seven annotators, and a psychology/linguistics expert made the final decision.} participated in the annotation process, all of whom are proficient Mandarin and English speakers (some speak more than three languages) and have at least a Master's degree. 
(2) \textbf{Drivelology Detection and Tagging.}~~Each annotator was tasked with determining whether a given text is non-Drivelology or Drivelology. 
Non-Drivelology includes both normal, meaningful sentences and pure nonsense that lacks rhetorical or semantic structure. 
If the text is identified as Drivelology, the annotators then perform a multi-label classification task, assigning one or more of the following categories to the sample: Misdirection, Paradox, Switchbait, Inversion, and Wordplay. 
(3) \textbf{Implicit Narrative Writing.}~~Given a Drivelology sample, we first prompt GPT-4 to generate narrative descriptions, illustrating the Drivelology's narrative and explaining the underlying meaning. 
Human annotators then double-check and modify the contents through dialogue interactions with the GPT-4 model to obtain a correct narrative. 
Additionally, we prompt GPT-4 to generate four hard negative counterparts to form a multiple-choice question answering task for our experiment. 
As narrative writing is inherently open-ended and involve subjectivity, we additionally frame this as selection tasks, and ensure that the correct option is clearly and objectively superior than the negative options to mitigate subjectivity. 
Following \citet{achiam2023gpt}, We primarily rely on human annotators to obtain gold-standard annotations, while allowing the annotators to collaborate with GPT-4. 
(4) \textbf{Quality Check with Verification.} To further minimise annotation errors, an experienced meta-reviewer with a background in linguistics and psychology systematically reviewed all annotated samples. The meta-reviewer excludes the samples with ambiguous or controversial narratives as some of them may introduce bias. This process ensures the quality of the annotated components for benchmark dataset construction. 


\begin{table*}[!t]
\centering
\resizebox{\linewidth}{!}{
\begin{tabular}{p{8cm} p{8cm} p{6cm}}
\toprule
\textbf{Text} & \textbf{Translated Text} & \textbf{Taggings} \\
\midrule
\begin{CJK*}{UTF8}{bkai}夜店這種地方還是少去，耳朵會聾掉。我陪朋友去過一次，後來男友叫我不要去，我都聽不見。\end{CJK*} &Nightclubs are the kind of place you should go to less, your ears will go deaf. I went once to accompany a friend, and later my boyfriend told me not to go, but I couldn’t hear him. &wordplay \\
\midrule
\begin{CJK*}{UTF8}{bkai}愛一個人是藏不住的，但愛兩個一定要藏住。\end{CJK*} &Loving someone cannot be hidden, but loving two people must be hidden. &switchbait \\
\midrule
\begin{CJK*}{UTF8}{bkai}母親節已經想好要送什麼了。給自己買件新衣服，送媽媽一個漂亮的女兒。\end{CJK*} &Mother's Day gift is already decided. Buy myself a new dress and give my mom a beautiful daughter. &misdirection \\
\midrule
\begin{CJK*}{UTF8}{bkai}同學：你都怎麼作弊？明天段考。 我：偷偷的把課本的內容都記在腦袋裡，老師根本抓不到。\end{CJK*} &Classmate: How do you usually cheat? The midterm is tomorrow. Me: I secretly memorize all the contents of the textbook in my head, the teacher can't catch me at all. &inversion \\
\midrule
\begin{CJK*}{UTF8}{gbsn}只要夫妻两个人互相信任，四个人就能相安无事。\end{CJK*} &As long as the husband and wife trust each other, four people can get along in peace. &inversion, wordplay \\
\midrule
\begin{CJK*}{UTF8}{bkai}以前我老婆對我真的超兇的，後來我就讓他去學空手道跟劍道。至少現在他打我之前，會先跟我鞠躬。\end{CJK*} &In the past, my wife was really super mean to me. Later, I let her go learn karate and kendo. At least now, before she hits me, she will bow to me first. &inversion, switchbait \\
\midrule
\begin{CJK*}{UTF8}{gbsn}高速公路旁的警语写着：开车请看前方。\end{CJK*} &The warning sign by the highway reads: Please keep your eyes on the road while driving. &inversion, paradox \\
\midrule
\begin{CJK*}{UTF8}{gbsn}女孩从不会在意你开什么颜色的法拉利。\end{CJK*} &A girl will never care what color Ferrari you drive. &misdirection, inversion, wordplay \\
\midrule
\begin{CJK*}{UTF8}{bkai}學生：老師，我媽要我問一下我的成績出來了嗎？老師：你等一下。學生：好的。老師：09 55 34 20 47。學生：打不通。老師：這是成績。\end{CJK*} & Student: Teacher, my mom asked me to check if my grades are out yet? Teacher: Just a moment. Student: Okay. Teacher: 09 55 34 20 47. Student: Can't get through. Teacher: That's your grade. &misdirection \\
\midrule
\begin{CJK*}{UTF8}{bkai}我：今年過年我要帶女朋友回去喔。老媽：幾歲，哪裡人？我：到時候你們自己問他，他很溫柔可愛體貼，沒什麼缺點。老媽：你就是他最大的缺點。\end{CJK*} &Me: This year during Lunar New Year, I’m bringing my girlfriend home. Mom: How old is she? Where is she from? Me: You can ask her yourself then. She’s gentle, cute, thoughtful—she doesn’t have many flaws. Mom: You’re her biggest flaw. &misdirection \\
\midrule
\begin{CJK*}{UTF8}{min}私の長所は素直に間違いを認めることです。 短所は、決して間違いを改めないことです。\end{CJK*} &My strength is that I can honestly admit my mistakes. My weakness is that I never correct my mistakes. &paradox \\
\midrule
\begin{CJK*}{UTF8}{min}お客様のおかげで忍耐力がアップしてきました。\end{CJK*} &Thanks to the customer, my patience has improved. &inversion, wordplay \\
\midrule
\begin{CJK*}{UTF8}{mj}제가 못하는 것 빼고는 다 잘해요\end{CJK*} &I'm good at everything except what I can't do. &paradox\\
\midrule
\begin{CJK*}{UTF8}{mj}A: 돌잔치 결혼 장례식 등등 한달전에 얘기하셈 연차 올려야하니 B: 한달전은 빡세네 장례식 한달전 예고면 살인아님?\end{CJK*} &A: Let me know a month in advance for events like funerals. B: A month's notice for a funeral? That’s premeditated murder! &paradox \\
\midrule
\begin{CJK*}{UTF8}{mj}여자친구：나 살찐 거 같아？남자친구：넌 살이 문제가 아니야。\end{CJK*} &Girlfriend: Do you think I gained weight? Boyfriend: Your problem isn’t your weight. &inversion \\
\midrule
\begin{CJK*}{UTF8}{mj}집에 불이 났다. 온 가족이 당황해서 소리친다. 아버지 : 야, 119가 몇 번이야? 119가 몇 번이냐고 !!!아들: 아버지, 이럴 때일수록 침착하셔야 돼요. 제가 114에 전화해서 물어볼게요 ~~ \end{CJK*} &The house caught fire. The whole family was panicking and shouting. Father: Hey, what's the number for 119? What's the number for 119!!! Son: Dad, you need to stay calm in situations like this. I'll call 114 and ask. &misdirection, switchbait \\
\midrule
Quel est le coquillage le plus léger? La palourde. & What is the lightest shell? The clam. &wordplay \\
\midrule
Une vague amoureuse du vent lui demande : Est-ce que tu peux me faire une petite bise aujourd’hui? &A wave, in love with the wind, asks: ``Can you give me a little kiss today?'' &wordplay \\
\midrule
Que horrible cuando tu mamá te da instrucciones y tú estás medio dormida, entonces no te acuerdas si tenías que lavar la basura, colgar al perro o sacar a pasear la ropa. &How awful when your mom gives you instructions while you're half asleep, so you can't remember whether you were supposed to wash the trash, walk the dog, or take the clothes out for a walk. &misdirection \\
\midrule
C’est l’histoire de deux pommes de terre. Une d’elles se fait écraser et l’autre s’écrie: Oh purée ! &It's the story of two potatoes. One of them gets crushed, and the other exclaims: Oh mashed potatoes! &wordplay \\
\midrule
Pourquoi est-ce que Hulk a un beau jardin? Parce qu’il a la main verte. &Why does Hulk have a beautiful garden? Because he has a green thumb. &switchbait, wordplay \\
\midrule
Que fait un employé de chez Sephora à sa pause clope ? Il parfumer. &What does a Sephora employee do during their cigarette break? They perfume. &switchbait, wordplay \\
\midrule
Qu'est ce qu'une lampe moche ? Un LEDron. &What do you call an ugly lamp? A LED-boring. &wordplay \\
\midrule
Pourquoi est ce que Potter est triste ? Parce que personne Harry à sa blague. & Why is Potter sad? Because no one Harry gets his joke. &switchbait, wordplay \\
\bottomrule
\end{tabular}
}
\caption{Representative examples of Drivelology.}
\label{tab:drivelology_examples}
\end{table*}

\subsection{Dataset Distribution}

\begin{table}[!t]
\centering
\resizebox{\linewidth}{!}{
\begin{tabular}{lccc}\toprule
\textbf{Language} &\textbf{Drivelology} &\textbf{Non-Drivelology} &\textbf{Total} \\\midrule
Mandarin &277 &194 &471 \\
English &93 &75 &168 \\
Spanish &69 &68 &137 \\
French &62 &80 &142 \\
Korean &52 &92 &144 \\
Japanese &47 &91 &138 \\
\midrule
Total &600 &600 &1200 \\
\bottomrule
\end{tabular}
}
\caption{Language distribution of Drivelology and non-Drivelology samples in the \textsc{DrivelHub}. 
Mandarin includes Simplified Chinese and Traditional Chinese.
}\label{tab:distribution}
\end{table}

\begin{figure*}[!t]
    \centering
    \includegraphics[width=0.9\textwidth]{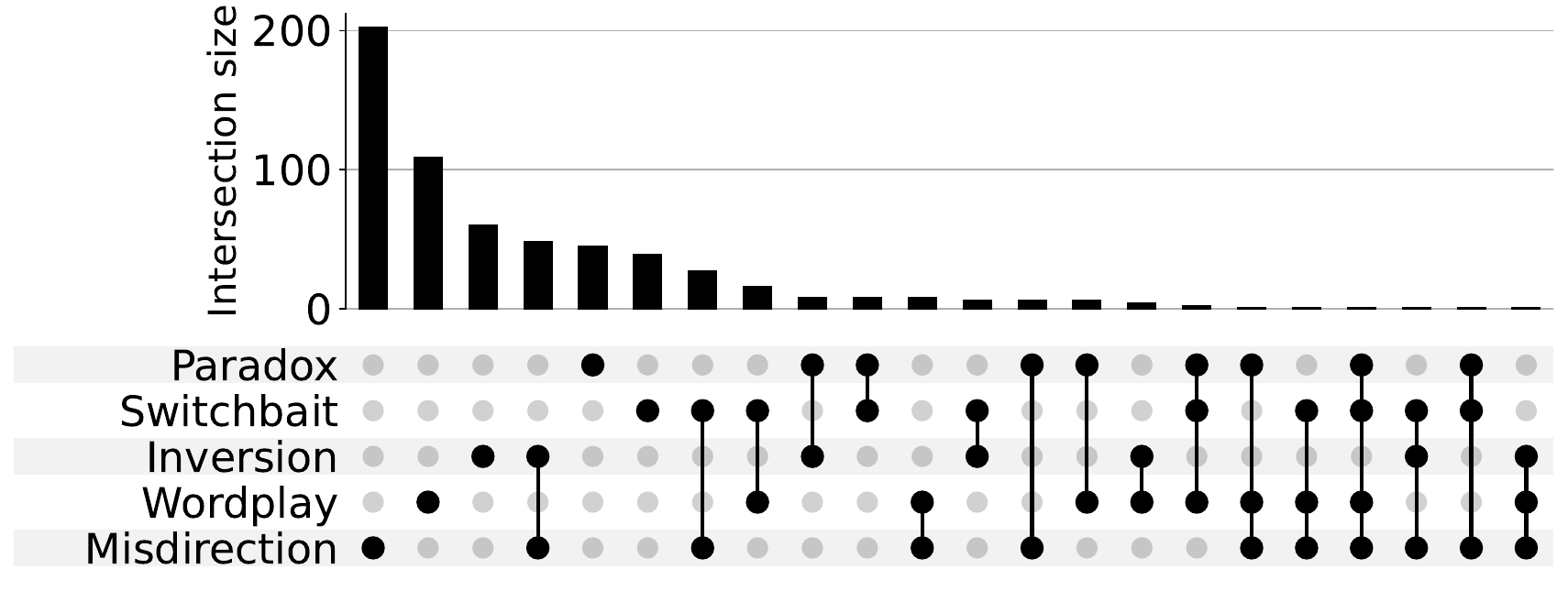}
    \caption{UpSet plot \cite{lex2014upset} illustrating the overlap and intersection sizes among Drivelology categories. Each vertical bar represents the number of samples belonging to a specific combination of categories, as indicated by the connected black dots below. Categories include Misdirection, Paradox, Switchbait, Inversion, and Wordplay.}
    \label{fig:label_distribution}
\end{figure*}

Table~\ref{tab:distribution} presents the language distribution of samples in the \textsc{DrivelHub} dataset. 
As shown, the dataset is skewed toward Mandarin, which accounts for 277 out of the total Drivelology samples. 
In contrast, other languages such as Japanese and Korean are present only in limited quantities. 
To characterise the distribution and overlap of annotation categories in our dataset, we present an UpSet plot \cite{lex2014upset} in Figure~~\ref{fig:label_distribution}, summarising intersections among the five Drivelology categories.

\section{Experimental Details}

\subsection{Experiment Prompts}
\label{sec:experiment_prompts}

To ensure reproducibility and transparency, we provide the exact prompts used in each of our experimental tasks. 
These prompts were carefully designed to probe different aspects of Drivelology comprehension and generation across various LLMs. 
Below, we detail the prompts for each task: Drivelology Detection (Figure~\ref{fig:prompt_detection}), Drivelology Tagging (Figure~\ref{fig:prompt_tagging}), Narrative Writing (Figure~\ref{fig:prompt_writing} for generation and Figure~\ref{fig:prompt_writing_evaluation} for evaluation), Narrative Selection (Figure~\ref{fig:prompt_easy_selection} for Easy and Figure~\ref{fig:prompt_hard_selection} for Hard).

\begin{table*}[!t]
\centering
\begin{tabular}{lcrrrrrr}
\toprule
\multirow{2}{*}{\textbf{Prompt}} & \multirow{2}{*}{\textbf{Model Size}} & \multicolumn{2}{c}{\textbf{Narrative}} & \multicolumn{2}{c}{\textbf{MCQA}} & \multicolumn{2}{c}{\textbf{Classification}} \\
\cmidrule(lr){3-4} \cmidrule(lr){5-6} \cmidrule(lr){7-8}
& & \textbf{BERT} & \textbf{GPT} & \textbf{Easy} & \textbf{Hard} & \textbf{Detect} & \textbf{Tag} \\
\midrule
\multirow{3}{*}{English} & 4B & 77.45 & 2.64 & 81.00 & 6.00 & 66.80 & 43.21 \\
& 8B & 85.91 & 2.64 & 83.17 & 26.78 & 65.00 & 38.04 \\
& 14B & 86.00 & 2.67 & 83.96 & 46.66 & 73.57 & 45.19 \\
\midrule
\multirow{3}{*}{Mandarin} & 4B & 67.79 & 2.96 & 77.61 & 2.44 & 62.86 & 46.10 \\
& 8B & 65.07 & 3.08 & 81.11 & 19.11 & 78.81 & 41.71 \\
& 14B & 64.23 & 3.19 & 82.56 & 51.69 & 77.62 & 49.35 \\
\bottomrule
\end{tabular}
\caption{Performance of Qwen3 models of varying sizes (4B, 8B, 14B) across different tasks.}
\label{tab:qwen_scaling_full}
\end{table*}



\section{Drivelology Generation}
\label{sec:drivelology_generation}

To explore the generative capabilities of LLMs in Drivelology, we conducted a case study using GPT-4. 
Our goal was to assess whether the model can produce contextually natural and pragmatically rich Drivelology examples, focusing on both surface form and deeper pragmatic alignment.

We designed two experimental settings: (1) Minimal Guidance and (2) Guidance With Category Definitions. 
In the Minimal Guidance setting, the model received only a brief introduction to Drivelology, without any example texts or category definitions. 
In the Category Definitions setting, the model was provided with detailed definitions of the five Drivelology categories (see \S\ref{sec:drivelology}). 
For each stage, we tested three prompting strategies: zero-shot, one-shot, and five-shot. 

\subsection{Findings}

Our investigation into GPT-4's generative capabilities reveals a significant gap between mimicking linguistic forms and achieving genuine pragmatic depth. 

\noindent\textbf{Minimal Guidance.}~~When prompted with only a brief description of Drivelology, GPT-4 relied heavily on surface-level cues such as \textit{paradoxical}, \textit{unexpected twist}, or \textit{nonsensical}. 
The resulting outputs were typically simple, declarative statements containing superficial contradictions (e.g., ``He's an honest liar'' or ``I bought a one-way ticket with unlimited uses''). 
These examples mimicked the form of Drivelology but lacked the semantic depth, layered meaning, and interpretive tension that define the genre.

\noindent\textbf{With Category Definitions.}~~Providing explicit category definitions led to more complex outputs, including richer character interactions, emotional cues, and linguistic characteristics like personification. 
For example: ``He says he's vegetarian, but only eats plants that scream -- like carrots that wail when pulled from the ground.'' While this sentence demonstrates greater creativity and engagement with paradox and wordplay, it still falls short of Drivelology's essential qualities. The supposed contradiction is not inherently ironic, and the interpretive tension remains weak. 
Additionally, increasing the number of examples (five-shot) did not improve output quality. 
Instead, it often exposed deeper structural and semantic issues. Some outputs suffered from syntactic misalignment (e.g., ``It's not that you can't love others, it's that love can't you,'' which is ungrammatical and uninterpretable), while others exhibited shallow or logically incompatible contradictions (e.g., ``It's not that I don't want to work hard, it's that I've worked so hard it looks like I'm not trying'').

Across all settings, GPT-4 struggled to internalise the subtle requirements of Drivelology. 
Out of 20 generations prompted with examples, only one output achieved comprehensive alignment between topic, rhetorical category, and sentence structure. 
For example: ``\begin{CJK*}{UTF8}{gkai}这本书太深奥了，我花了一整晚没看懂封面。\end{CJK*}(This book is too profound, I spent the whole night and still couldn’t understand the cover).''
Although providing more examples led to slightly more complex narratives, the outputs consistently lacked Drivelology's hallmark features: contextual misdirection, interpretive layering, and rhetorical paradox. 
These shortcomings were especially pronounced in scenarios requiring cultural knowledge, emotional nuance, and inferential reasoning. 
Overall, our findings highlight the persistent challenges LLMs face in generating text that aligns with the deeper pragmatic and rhetorical demands of Drivelology.

\section{Future Work}

While this work successfully introduces the \textsc{DrivelHub} dataset and benchmarks the limitations of current LLMs, the rich structure of our data opens up significant avenues for future research. We outline two key directions: advancing model training methodologies and developing a robust framework for evaluating Drivelology generation.

\subsection{Advancing Model Training with the MCQA Task}

We have identified that the Narrative Selection (MCQA) task within our dataset is a perfect fit for GRPO \cite{deepseek-math}. GRPO is an advanced preference optimisation technique that allows a model to learn from the relative preferences within a group of candidate responses, rather than relying on simple pairwise \cite{rafailov2023direct} or scalar rewards \cite{schulman2017proximal}. By learning from a full ranking of multiple candidates, the model receives a much richer training signal. 
The design of our MCQA task naturally lends itself to this paradigm. For each Drivelology sample, we provide one correct narrative and several carefully crafted distractors. This setup creates an inherent group-wise ranking (i.e., the correct option is preferred over all incorrect options), which can be directly leveraged by GRPO. Future work should explore fine-tuning LLMs using GRPO on the \textsc{DrivelHub} MCQA data. We hypothesise that this approach could substantially improve a model's ability to discern subtle semantic and pragmatic distinctions, thereby enhancing its capacity for the deep, non-linear reasoning required to truly comprehend Drivelology. This would represent a significant step toward closing the gap between statistical fluency and genuine cognitive understanding that our current work highlights.

\subsection{Developing Metrics for Drivelology Generation}

Our current study focuses primarily on the understanding and reasoning abilities of LLMs rather than their capacity for generation. A significant area for future work is to establish a comprehensive framework for evaluating generated Drivelology. A key limitation to address is the absence of metrics capable of quantifying the qualities that make a text Drivelological. 
Simply measuring fluency or grammatical correctness is insufficient. A robust evaluation would require developing novel metrics to assess specific aspects of a generated Drivelology text, such as: (1) \textbf{Entertainability}: How humorous, witty, or engaging is the output? (2) \textbf{Cohesion and Paradoxical Depth}: How well does the output maintain surface-level coherence while simultaneously embedding a meaningful, non-obvious contradiction or twist? (3) \textbf{Originality}: How surprising or non-formulaic is the output? Does it avoid simply rehashing common Drivelological text or existing templates? (4) \textbf{Cultural Resonance}: How well does the output tap into shared cultural knowledge, social scripts, or contemporary memes to create its meaning?

Furthermore, a robust framework must test for \textit{controllable generation} -- the ability to create Drivelology that meets specific constraints, like producing an ``inversion'' about ``technology.'' Success here would be a strong signal of true comprehension. While developing this framework is challenging, it is essential for two reasons: it would allow for more rigorous model assessment and provide clearer targets for training. This creates a powerful feedback loop where better evaluation drives better generation, which in turn deepens the model's core reasoning, ultimately leading to LLMs that can truly master ``nonsense with depth.''

\clearpage

\begin{figure*}[htbp]
    \def\arraystretch{1.5}
	\fontsize{9}{10}\selectfont
     \hspace{-2mm}
	\setlength{\tabcolsep}{0.8mm}
	\centering
	\begin{tabular}{p{130mm}}
	\toprule
\textbf{Human Guidelines:}\\
\hdashline
\# Annotation Guidelines for Drivelology Dataset
\\
\#\# Introduction
\\
These guidelines are designed to assist annotators in accurately labelling samples for the Drivelology dataset. Annotators should familiarise themselves with the definitions and characteristics of Drivelology and non-Drivelology texts before proceeding.
\\
\#\# Definitions

\begin{itemize}
    \item Drivelology:
    \begin{itemize}
        \item Description: Statements that appear logically coherent but contain deeper, often paradoxical meanings. These challenge conventional interpretation by blending surface-level nonsense with underlying depth, often incorporating elements of humour, irony, or sarcasm. Understanding Drivelology requires contextual insight and emotional interpretation.
        \item Examples:
        \begin{itemize}
            \item ``I bought a book on how to solve 50\% of my problems, so I bought two books.''
            \item ``Loving someone cannot be hidden, but loving two people must be hidden.''
        \end{itemize}
    \end{itemize}
    \item non-Drivelology:
    \begin{itemize}
        \item Description: This includes pure nonsense (grammatically correct but semantically meaningless statements) and normal sentences, including quotes or proverbs, that convey clear or straightforward information without the layered complexity characteristic of Drivelology.
        \item Examples:
        \begin{itemize}
            \item ``The cat sat on the mat.'' (normal sentence)
            \item ``Colourless green ideas sleep furiously.'' (pure nonsense)
        \end{itemize}
    \end{itemize}
\end{itemize}
\\
\#\# Annotation Tasks
\begin{itemize}
    \item Drivelology Tagging
    \begin{itemize}
        \item Task: Classify Drivelology samples into one or more categories only if the sample is Drivelology:
        \begin{itemize}
            \item Misdirection: A rhetorical technique where the focus shifts but connects back to the original topic through indirect hints.
            \item Paradox: A statement that combines ideas that do not logically fit together but conveys a deeper meaning.
            \item Switchbait: A language trick that changes meaning based on cultural knowledge or idioms.
            \item Inversion: Rearranging the usual order of words or ideas to create a surprising effect.
            \item Wordplay: Creative use of language through puns or double meanings.
        \end{itemize}
        \item Instructions:
        \begin{itemize}
            \item Identify the primary characteristics (i.e., the first strong impression) of the text.
            \item Assign one or more categories based on the definitions above.
        \end{itemize}
    \end{itemize}
    \item Implicit Narrative Writing
    \begin{itemize}
        \item Task: Generate a detailed description illustrating the implicit narrative of the Drivelology text.
        \item Instructions:
        \begin{itemize}
            \item Analyse the text to uncover underlying themes, messages, or emotional undertones.
            \item Write a narrative that reflects the deeper significance of the text, going beyond a surface-level summary.
            \item Generate four contextualised, plausible, but ultimately incorrect narrative, wrong understanding of the given Drivelology text, each within three sentences as distractors. Keep the length and style the same as the correct narrative, and keep these negative narratives difficult to tell from the positive narrative.
        \end{itemize}
    \end{itemize}
\end{itemize}
\\
\#\# Quality Assurance
\\
Each annotation will undergo a review process where a meta-reviewer will assess the annotations for consistency and accuracy. 
Annotators should mark down any samples that exhibit ambiguities or uncertainties during the annotation process. 
The meta-reviewer will review these marked samples and finalise the answer based on a thorough evaluation.
\\
\bottomrule
\end{tabular}
\caption{Guidelines for human annotators.} 
\label{fig:human_guidelines}
\vspace{2mm}
\end{figure*}

\clearpage

\begin{figure*}[htbp]
    \def\arraystretch{1.5}
	\fontsize{9}{10}\selectfont
     \hspace{-2mm}
	\setlength{\tabcolsep}{0.8mm}
	\centering
	\begin{tabular}{p{130mm}}
	\toprule
\textbf{Prompt1:}\\
\hdashline
Instruction:

Classify whether the given text is a Drivelology sample or not.
\\
Definition:

 - Drivelology: Statements that appear logically coherent but contain deeper, often paradoxical meanings. These challenge conventional interpretation by blending surface-level nonsense with underlying depth, often incorporating elements of humor, irony, or sarcasm, and requiring contextual understanding and emotional insight to unravel their true significance.
 
 - non-Drivelology: This includes pure nonsense (grammatically correct but semantically meaningless statements, such as "boys will be boys") and normal sentences, including quotes or proverbs, that convey clear or straightforward information without the layered complexity characteristic of Drivelology.
\\
Input Text:

\{text\}
\\
Output Format:

Please provide the output in JSON format with the following keys:

 - answer: Specify whether the text is "Drivelology" or "non-Drivelology."

 - reason: Provide a clear explanation of why the text is classified as Drivelology or not.
\\
\textbf{Prompt2:}\\
\hdashline
Instruction:

Classify whether the given text is a Drivelology sample or not.
\\
Definitions:

 - Drivelology: Statements that appear logically coherent but contain deeper, often paradoxical meanings. These challenge conventional interpretation by blending surface-level nonsense with underlying depth, often incorporating elements of humor, irony, or sarcasm, and requiring contextual understanding and emotional insight to unravel their true significance.
 
 - non-Drivelology: This includes pure nonsense (grammatically correct but semantically meaningless statements) and normal sentences, including quotes or proverbs, that convey clear or straightforward information without the layered complexity characteristic of Drivelology.
\\
Input Text:

\{text\}
\\
Instructions for Reasoning:

Analyse the input text by comparing it to the definitions above. Identify whether it contains logical coherence, paradox, layered meaning, or requires emotional/contextual insight. If uncertain, select the category that best fits and explain your reasoning.
\\
Output Format:

Please provide the output in JSON format with the following keys:

 - answer: Specify "Drivelology" or "non-Drivelology."
 
 - reason: Clearly explain why the text was classified as such, referencing specific features from the definitions.
\\
\textbf{Prompt3:}\\
\hdashline
Classify the text as "Drivelology" or "non-Drivelology."
\\
Definitions:

 - Drivelology: Logically coherent statements with paradox, layered or hidden meaning, often using humor, irony, or sarcasm. These require emotional or contextual insight to interpret.
 
 - non-Drivelology: Pure nonsense or straightforward statements without hidden complexity.
\\
Text:

\{text\}
\\
Reasoning:

Decide based on the definitions above. If uncertain, choose the closest fit and briefly explain.

Output (JSON only):

\{\\
\quad\quad"answer": "Drivelology",
  
\quad\quad"reason": "The text contains underlying meaning, fitting the Drivelology definition."\\
\}
\\
\bottomrule
\end{tabular}
\caption{Prompts for Drivelology Detection task. 
} 
\label{fig:prompt_detection}
\vspace{2mm}
\end{figure*}

\clearpage

\begin{figure*}[htbp]
    \def\arraystretch{1.5}
	\fontsize{9}{10}\selectfont
     \hspace{-2mm}
	\setlength{\tabcolsep}{0.8mm}
	\centering
	\begin{tabular}{p{130mm}}
	\toprule
\textbf{Prompt1:}\\
\hdashline
Instruction:

Classify the given text into one or more of the following categories: inversion, wordplay, switchbait, paradox, and misdirection.
\\
Definitions:

 - inversion: INVERSION DEFINITION.
 
 - wordplay: WORDPLAY DEFINITION.
 
 - switchbait: WITCHBAIT DEFINITION.
 
 - paradox: PARADOX DEFINITION. 
 
 - misdirection: MISDIRECTION DEFINITION. 
\\
Input Text:

\{text\}
\\
Output Format:

Please provide the output in JSON format with the following keys:

 - answer: List the applicable comma-separated categories for the text (e.g., "category1, category2").
 
 - reason: Provide a clear explanation for why the text is classified into each category.
\\
\textbf{Prompt2:}\\
\hdashline
Instruction:

Analyse the input text and classify it into one or more of the following categories: inversion, wordplay, switchbait, paradox, and misdirection. 
Use the definitions below to guide your classification.
\\
Definitions:

 - inversion: INVERSION DEFINITION.
 
 - wordplay: WORDPLAY DEFINITION.
 
 - switchbait: WITCHBAIT DEFINITION.
 
 - paradox: PARADOX DEFINITION. 
 
 - misdirection: MISDIRECTION DEFINITION. 
\\
Input Text:

\{text\}
\\
Output Format (JSON):

\{\\
\quad\quad"answer": "category1, category2, ...",
  
\quad\quad"reason": "Explain why the text fits each chosen category based on the definitions."\\
\}
\\
\textbf{Prompt3:}\\
\hdashline
Instruction:

Examine the input text and determine which of the following categories it belongs to: inversion, wordplay, switchbait, paradox, and misdirection.  
Base your classification strictly on the definitions provided below.
\\
Definitions:

 - inversion: INVERSION DEFINITION.
 
 - wordplay: WORDPLAY DEFINITION.
 
 - switchbait: WITCHBAIT DEFINITION.
 
 - paradox: PARADOX DEFINITION. 
 
 - misdirection: MISDIRECTION DEFINITION. 
\\
Input Text:

\{text\}
\\
Please provide the output in JSON format: 

\{\\
\quad\quad"answer": "category1, category2, ...",
  
\quad\quad"reason": "Briefly explain how the text fits each selected category, using the definitions as a basis."\\
\}
\\
\bottomrule
\end{tabular}
\caption{Prompts for Drivelology Tagging task. 
} 
\label{fig:prompt_tagging}
\vspace{2mm}
\end{figure*}

\clearpage

\begin{figure*}[htbp]
    \def\arraystretch{1.5}
	\fontsize{9}{10}\selectfont
     \hspace{-2mm}
	\setlength{\tabcolsep}{0.8mm}
	\centering
	\begin{tabular}{p{130mm}}
	\toprule
\textbf{Prompt1:}\\
\hdashline
You need to first read and understand the text given. 
Generate a detailed description to illustrate the implicit narrative of the text.
\\
Text: \{text\}
\\
Output format should be JSON with the following keys:

 - narrative: The narrative of the text in English.
\\
\textbf{Prompt2:}\\
\hdashline
Read and understand the provided text carefully. 
\\
Task: Generate a detailed description that illustrates the implicit narrative of the text.
\\
Input Text: \{text\}
\\
Output Format:

\{\\
\quad"narrative": "The narrative of the text."\\
\}
\\
Please ensure the output is in JSON format and contains the key "narrative" with the developed description of the implicit narrative derived from the input text.
\\
\textbf{Prompt3:}\\
\hdashline
Read and understand the provided text carefully.
\\
Task:

Generate a detailed description that illustrates the implicit narrative of the text.
By "implicit narrative," we mean the underlying message, theme, perspective, or emotional undertone that is not directly stated but can be inferred from the text. 
Your description should go beyond surface-level summary and provide insights into the text's underlying themes, perspectives, or intentions.
\\
Input Text:

{text}
\\
Output Format:

Output only the JSON object, with no extra commentary or explanation.

\{\\
\quad"narrative": "A detailed description of the implicit narrative, including the underlying theme, emotional tone, and implied perspective."\\
\}
\\
\bottomrule
\end{tabular}
\caption{Prompts for Narrative Writing task. 
} 
\label{fig:prompt_writing}
\vspace{2mm}
\end{figure*}

\clearpage

\begin{figure*}[htbp]
    \def\arraystretch{1.5}
	\fontsize{9}{10}\selectfont
     \hspace{-2mm}
	\setlength{\tabcolsep}{0.8mm}
	\centering
	\begin{tabular}{p{130mm}}
	\toprule
\textbf{Prompt1:}\\
\hdashline
Task:
\\
Evaluate how accurately the candidate narrative matches the given reference narrative. 
\\
Use a scale from 1 to 5, where 1 indicates the least accuracy and 5 indicates the highest accuracy.
\\
 - Candidate Narrative: \{candidate\}
\\
 - Reference Narrative: \{reference\}
\\
Output Format:
\\
Please provide the output in JSON format with the following key:
\\
 - score: The score indicating the level of matching, ranging from 1 to 5.
\\
\bottomrule
\end{tabular}
\caption{Prompts for evaluating Narrative Writing task. 
} 
\label{fig:prompt_writing_evaluation}
\vspace{2mm}
\end{figure*}

\clearpage

\begin{figure*}[htbp]
    \def\arraystretch{1.5}
	\fontsize{9}{10}\selectfont
     \hspace{-2mm}
	\setlength{\tabcolsep}{0.8mm}
	\centering
	\begin{tabular}{p{130mm}}
	\toprule
\textbf{Prompt1:}\\
\hdashline
Tell me the best option in the following options which represents the underlying narrative of the text?
\\
Text: \{text\}
\\
A. \{narrative\_1\}

B. \{narrative\_2\}

C. \{narrative\_3\}

D. \{narrative\_4\}

E. \{narrative\_5\}
\\
Output format should be JSON with the following keys:

 - answer: The option the text belongs to, and it should be uppercase among A, B, C, D, E.
\\
\textbf{Prompt2:}\\
\hdashline
Tell me the best option from the list below that represents the underlying narrative of the text.
By "underlying narrative," we mean the main theme, implicit message, or perspective the text conveys.
\\
Text: \{text\}
\\
A. \{narrative\_1\}

B. \{narrative\_2\}

C. \{narrative\_3\}

D. \{narrative\_4\}

E. \{narrative\_5\}
\\
If more than one option seems plausible, pick the one that best represents the main narrative.
\\
Output format should be JSON with the following keys:

 - answer: The option should be a single uppercase letter among A, B, C, D, or E.
\\
Output only the JSON object, with no extra commentary.
\\
\textbf{Prompt3:}\\
\hdashline
Tell me which option best represents the underlying narrative (main theme, message, or perspective) of the text.
\\
Text: \{text\}
\\
A. \{narrative\_1\}

B. \{narrative\_2\}

C. \{narrative\_3\}

D. \{narrative\_4\}

E. \{narrative\_5\}
\\
If more than one fits, pick the best.
\\
Output only JSON:

 - answer: One uppercase letter: A, B, C, D, or E.
\\
Example:

\{\\
\quad"answer": "B"\\
\}
\\
\bottomrule
\end{tabular}
\caption{Prompts for Easy Narrative Selection task. 
} 
\label{fig:prompt_easy_selection}
\vspace{2mm}
\end{figure*}

\clearpage

\begin{figure*}[htbp]
    \def\arraystretch{1.5}
	\fontsize{9}{10}\selectfont
     \hspace{-2mm}
	\setlength{\tabcolsep}{0.8mm}
	\centering
	\begin{tabular}{p{130mm}}
	\toprule
\textbf{Prompt1:}\\
\hdashline
Tell me the best option in the following options which represents the underlying narrative of the text?
\\
Text: \{text\}
\\
A. \{narrative\_1\}

B. \{narrative\_2\}

C. \{narrative\_3\}

D. \{narrative\_4\}

E. None of the above.
\\
Output format should be JSON with the following keys:

 - answer: The option the text belongs to, and it should be uppercase among A, B, C, D, E.
\\
\textbf{Prompt2:}\\
\hdashline
Tell me the best option from the list below that represents the underlying narrative of the text.
By "underlying narrative," we mean the main theme, implicit message, or perspective the text conveys.
\\
Text: \{text\}
\\
A. \{narrative\_1\}

B. \{narrative\_2\}

C. \{narrative\_3\}

D. \{narrative\_4\}

E. None of the above.
\\
If none of the options fully fit, select "E. None of the above." If more than one option seems plausible, pick the one that best represents the main narrative.

Output format should be JSON with the following keys:

 - answer: The option the text belongs to, and it should be a single uppercase letter among A, B, C, D, or E.
\\
Output only the JSON object, with no extra commentary.
\\
Example output:

\{\\
\quad"answer": "B"\\
\}
\\
\textbf{Prompt3:}\\
\hdashline
Tell me which option best represents the underlying narrative (main theme, message, or perspective) of the text.
\\
Text: \{text\}
\\
A. \{narrative\_1\}

B. \{narrative\_2\}

C. \{narrative\_3\}

D. \{narrative\_4\}

E. None of the above.
\\
If none fit, choose E. If more than one fits, pick the best.
\\
Output only JSON:
 - answer: One uppercase letter: A, B, C, D, or E.
\\
Example:

\{\\
\quad"answer": "B"\\
\}
\\
\bottomrule
\end{tabular}
\caption{Prompts for Hard Narrative Selection task. 
} 
\label{fig:prompt_hard_selection}
\vspace{2mm}
\end{figure*}

\end{document}